\documentclass[a4paper]{svproc}
\usepackage{graphicx}
\DeclareGraphicsExtensions{.pdf,.png,.jpg}
\usepackage{url}
\usepackage{times}
\usepackage[numbers]{natbib}
\usepackage{ulem}
\usepackage{multicol}
\usepackage[bookmarks=true]{hyperref}
\usepackage{soul}
\usepackage{amsmath}
\usepackage{amsthm}
\usepackage{amsfonts}
\usepackage{enumitem}
\usepackage{tabularx}
\usepackage{marvosym}
\usepackage{cleveref}
\usepackage{wrapfig}
\DeclareMathOperator*{\argmin}{argmin} 
\usepackage{amssymb}
\usepackage{subcaption}
\usepackage[dvipsnames]{xcolor}
\usepackage{multirow}
\usepackage{nicematrix}
\usepackage{makecell}
\usepackage{booktabs}
\usepackage{algorithm}
\usepackage{algorithmic}
\usepackage{tikz}
\usetikzlibrary{calc}

\usepackage{blindtext,graphicx}
\usepackage[absolute]{textpos}
\usepackage{float}

\usepackage{setspace}

\newtheoremstyle{problemstyle}  % <name>
        {3pt}                                               % <space above>
        {3pt}                                               % <space below>
        {\normalfont}                               % <body font>
        {}                                                  % <indent amount}
        {\bfseries}                 % <theorem head font>
        {\normalfont:}         % <punctuation after theorem head>
        {.5em}                                          % <space after theorem head>
        {}                                                  % <theorem head spec (can be left empty, meaning `normal')>
\theoremstyle{problemstyle}

\makeatletter
\long\def\@makefntext#1{%
  \parindent 0pt% no hanging indent
  \noindent     % flush‐left
  #1            % the footnote text
}
\makeatother

\makeatletter
\newcommand\blfootnote[1]{%
  \begingroup
    \begin{NoHyper}%
      \renewcommand\thefootnote{}%
      \renewcommand\@makefnmark{}%
      \footnote{#1}%
      \addtocounter{footnote}{-1}%
    \end{NoHyper}%
  \endgroup
}
\makeatother
\Crefname{figure}{Fig.}{Figs.}

\begin{document}

\begin{textblock}{12}(4,1)
\noindent \centering \textit{Accepted for publication at the\\ International Conference of Experimental Robotics (ISER) 2025. \\ Final manuscript will be published as part of Springer Proceedings in Advanced Robotics (SPAR).}
\end{textblock}

\mainmatter 
% start of a contribution
\vspace{-3.0in}

\title{Real-time Remote Tracking and Autonomous Planning for Whale Rendezvous using Robots}

\titlerunning{Real-time Remote Tracking and Planning for Whale Rendezvous} 
\author{Sushmita Bhattacharya\inst{1,2(\text{\Letter})}\and
Ninad Jadhav\inst{1,2}\and
Hammad Izhar\inst{1,2}\and
Karen Li\inst{1,2}\and \\
Kevin George\inst{2}\and
Robert Wood\inst{1,2}\and
Stephanie Gil\inst{1,2}
}
\authorrunning{S. Bhattacharya \& N. Jadhav et al.}
\institute{
$^1$Harvard University, Cambridge, MA 02139, USA\\
\email{sushmita\_bhattacharya@g.harvard.edu} \\
$^2$Project CETI, New York, NY 10003, USA and Dominica
} 

\maketitle              % typeset the title of the contribution
    
\blfootnote{\scriptsize{S. Bhattacharya and N. Jadhav --- Equal contribution. Supplementary Material: https://harvard-react.github.io/ceti-avatars/}}
\blfootnote{\scriptsize{\textbf{Acknowledgment.} This study was funded by Project CETI grants from Dalio Philanthropies, Ocean X; Sea Grape Foundation; Virgin Unite, Rosamund Zander/Hansjorg Wyss (The Audacious Project: funding initiative by TED), and National Science Foundation (CAREER Grant No. CNS-2114733). We thank J. Siriska, D. Vogt, Z. Gonzalez-Peltier, Project CETI marine operations team, and the Government of Dominica.}}

\vspace{-0.425in}
\begin{abstract}
{
We introduce a system for real-time sperm whale rendezvous at sea using an autonomous uncrewed aerial vehicle. Our system employs model-based reinforcement learning that combines in situ sensor data with an empirical whale dive model to guide navigation decisions. Key challenges include (i) real-time acoustic tracking in the presence of multiple whales, (ii) distributed communication and decision-making for robot deployments, and (iii) on-board signal processing and long-range detection from fish-trackers. We evaluate our system by conducting rendezvous with sperm whales at sea in Dominica, performing hardware experiments on land, and running simulations using whale trajectories interpolated from marine biologists' surface observations.
}
\vspace{-0.125in}
\end{abstract}
\vspace{-0.4in}

\begin{figure*}[ht!]
    \centering
    \includegraphics[scale=0.145]{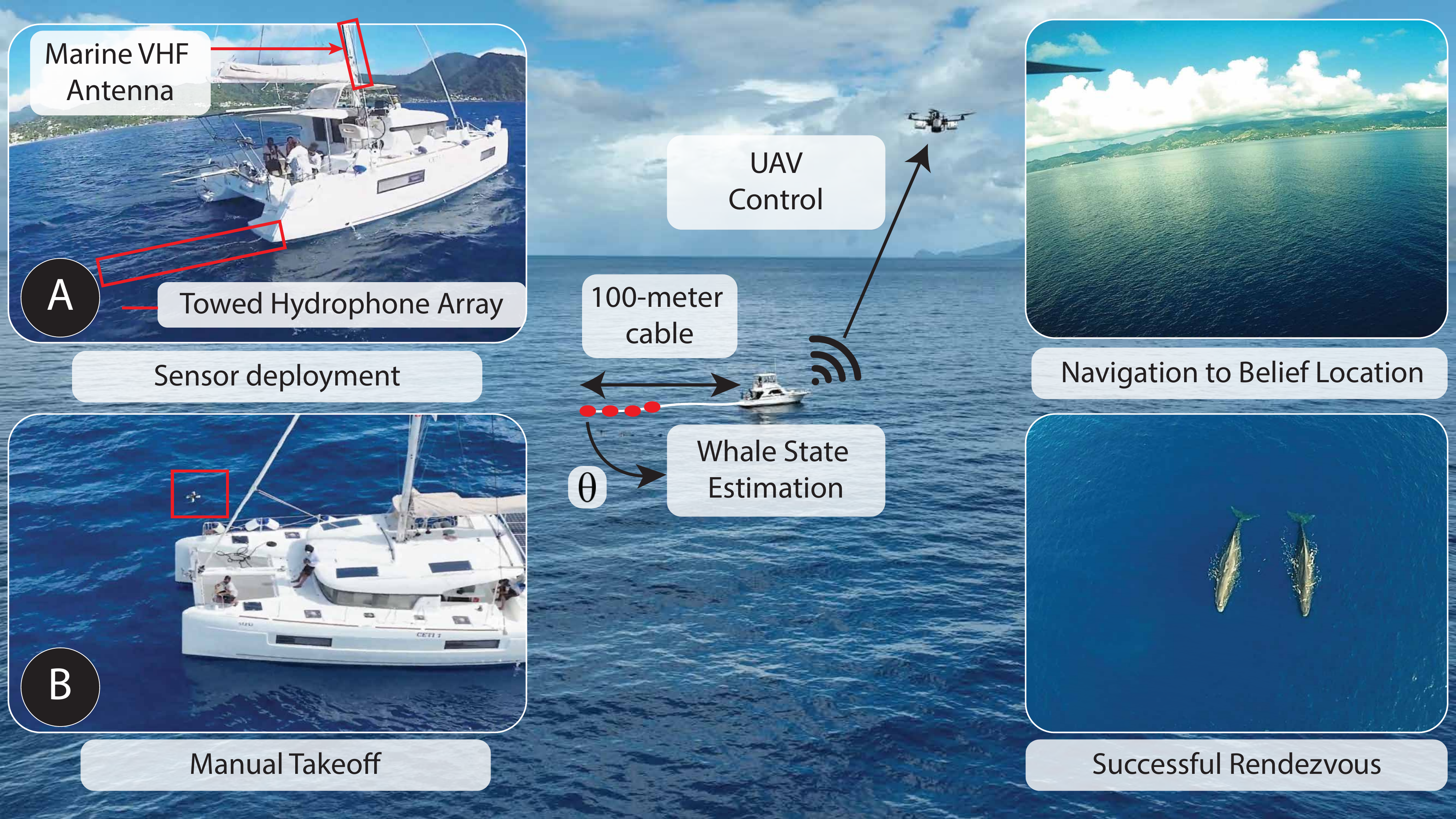}
\caption{\textbf{System setup for fielded
deployment in Dominica.} (A) The catamaran tows a linear hydrophone array to collect acoustic AOA measurements and has an antenna on its mast to detect VHF signals from an on-body fish-tracker. (B) The UAV is launched manually from the catamaran. (C) The UAV control module autonomously commands the UAV to a belief location. (D) Successful rendezvous with surfaced sperm whales.}
    \label{fig:intro_schematic}
\end{figure*}

\section{Introduction}
\label{introduction}
Deployment of robotic assets for close-range wildlife data collection can be tremendously useful~\cite{iScienceCETI,Chahine2025DecentralizedVA, Cliff2018RoboticET}. Timely positioning of such assets at the right place enables opportunistic sensing in environments where manual operation is labor-intensive and installing permanent infrastructure is impractical, such as at sea. Sperm whales offer a compelling and challenging test case for this capability: they surface only briefly, around ten minutes per hour~\cite{gero2014behavior}, and are typically tracked using noisy angle-of-arrival (AOA) estimates to their vocalizations~\cite{Premus2022}. Thus, real-time rendezvous requires robotic systems to operate under uncertainty, make sequential decisions, and coordinate remote sensing with sensor mobility to capture time-critical opportunities. We formulate rendezvous as a sequential decision-making problem and present a system that achieves rendezvous using autonomous robots and real-time, in-situ sensory input. 

Previous work in AOA-only target tracking focuses on using multiple sensors simultaneously or using additional sensing modalities, including visual sensing~\cite{karanam2018magnitude,Ning_ijrr_2024,10491136,zhou2011multirobot}. Our prior work developed an algorithmic framework that used post-processed non-overlapping AOA measurements collected from real whales in Dominica with a reinforcement learning (RL) based approach to demonstrate multi-whale rendezvous using a simulated robot team~\cite{scirob24}. Here, we extend this framework into a system (\Cref{fig:intro_schematic}) that integrates in situ data for (i) real-time whale state estimation (location and surfacing event) and (ii) autonomous control of an uncrewed aerial vehicle (UAV). Critical advancements from our previous work needed to attain real-time deployment include: (a) localization in the presence of overlapping acoustic AOA from simultaneously vocalizing whales, (b) addressing directional ambiguity of acoustic AOA where signals from the left and right appear identical due to the use of a linear towed hydrophone array, (c) beyond visual line-of-sight autonomous UAV control, and (d) demonstrating long-range very high frequency (VHF) signal detection and AOA computation to off-the-shelf VHF fish-trackers. GPS-enabled tags allow localization at the surface, but VHF trackers are preferred due to their longer battery life and power efficiency~\cite{Thomas2011WildlifeTT}.

% HI - Removes the constant use of "We use..." in the following commented out paragraph.
Our Dominica field system uses acoustic AOA to vocalizing whales computed by a linear hydrophone array towed by a manually crewed catamaran and then relays waypoint commands to a UAV.  A Gaussian mixture model (GMM) separates AOA measurements from distinct whale groups, and a particle filter estimates their positions. Directional ambiguity is resolved by maneuvering the catamaran to reposition the hydrophone array, which helps eliminate unlikely belief particles. By integrating a whale dive model into a model-based RL framework, the system generates UAV controls that respect realistic flight-time constraints. Long-range VHF reception can improve state estimation of tagged whales during silent surface intervals.

In proof-of-concept experiments at sea in Dominica, we achieve autonomous flights up to 1 km, rendezvousing within 200 meters of an untagged, surfaced whale on two occasions. We developed a lightweight VHF payload and demonstrated detection from an on-animal tag as a surfacing cue, with reliable reception at ranges up to 2 km in another experiment. Land-based tests showed on-board VHF AOA estimation with a mean angular error of 2.27° at an average range of 60 m. Complementary simulations evaluated capabilities not fully exercised at sea, including autonomous sensor maneuvers and takeoff decisions.

\begin{figure}[t]
    \centering
    \includegraphics[scale=0.195]{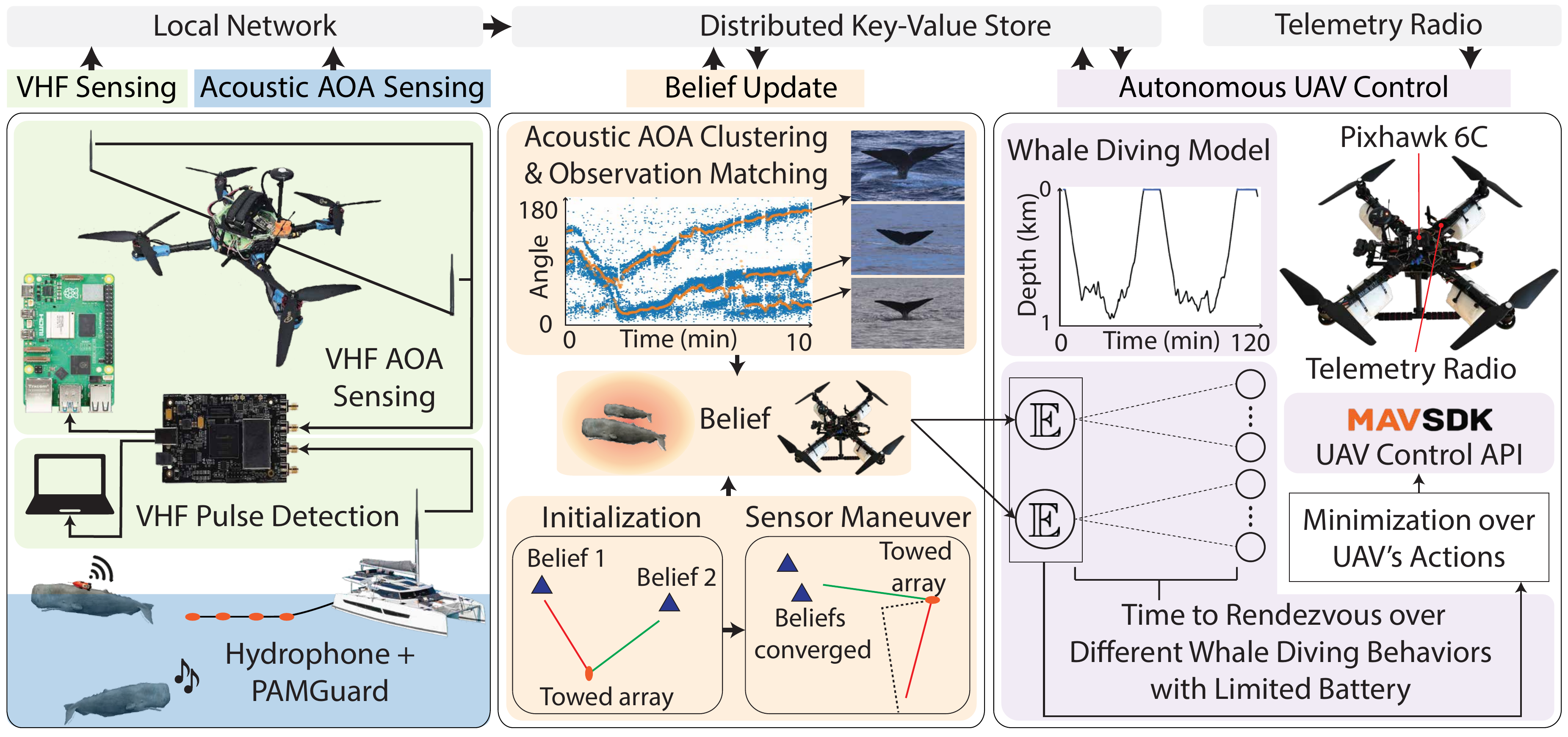}
    \vspace{-0.25in}
    \caption{\textbf{System architecture.} We use a hydrophone array and PAMGuard \cite{pamguard2008} to collect acoustic AOA measurements. A local network communicates sensor information to the state estimation module and a distributed key-value store handles synchronization. Our VHF sensing payload, deployed on the catamaran and a UAV, enables long range VHF signal detection and AOA estimation. We implement a model-based RL approach that integrates sensory information with a whale dive model~\cite{dswp} to control a UAV.
    }
   \label{fig:system_architecture}
   \vspace{-0.15in}
\end{figure}

\section{Technical Approach} \label{approach}
% In situ deployment requires integrating multiple sensing modalities, decision-making to better predict and respond to events based on real-time sensing, and UAV deployment for autonomous flight. We design a system  (\Cref{fig:system_architecture}) with distributed communication between the three modules to achieve real-time deployment.
We design a system (\Cref{fig:system_architecture}) that combines multiple sensing modalities, real-time decision-making, autonomous UAV flight, and distributed communication across different modules to enable real-time in situ sperm whale rendezvous.

\subsection{Sensing Module}
\subsubsection{Acoustic Sensing.}
We use a commercial linear hydrophone array with four elements, towed 100 meters behind a catamaran.
% We use a commercial linear hydrophone array consisting of four hydrophones, towed by a catamaran via a 100-meter cable. 
PAMGuard \cite{pamguard2008} processes whale vocalizations in real time to compute acoustic angle-of-arrival (AOA) estimates, which are batched and sent every 10 seconds to the state estimation module over a local network.
% PAMGuard \cite{pamguard2008} processes incoming whale vocalizations and computes real-time acoustic AOA estimates. These estimates are batched and transmitted every ten seconds over a local network to the state estimation module. 
The linear structure of the array induces a \textit{directional ambiguity} in the computed AOA, restricting the range of the estimates from $[0, 2\pi)$ to $[0, \pi)$. 
The array can detect signals from multiple sources, but PAMGuard does not differentiate between them.
% The array also detects signals from multiple sources simultaneously, but PAMGuard does not assign identifiers to distinguish them. 
When multiple whale groups are present,
% Thus, in the presence of two or more whale groups in spatially distinct locations, 
we must separate AOA estimates from each spatially distinct group to localize them individually. We refer to this as the \textit{group separation} problem—a relaxation of the source separation problem \cite{Diamant2025}, which aims to associate signals with individual whales based on vocal characteristics rather than spatial proximity.

\subsubsection{VHF Sensing.}
\begin{figure*}[t]
  % \vspace{-0.2in}
  \centering
  \includegraphics[scale=0.16]{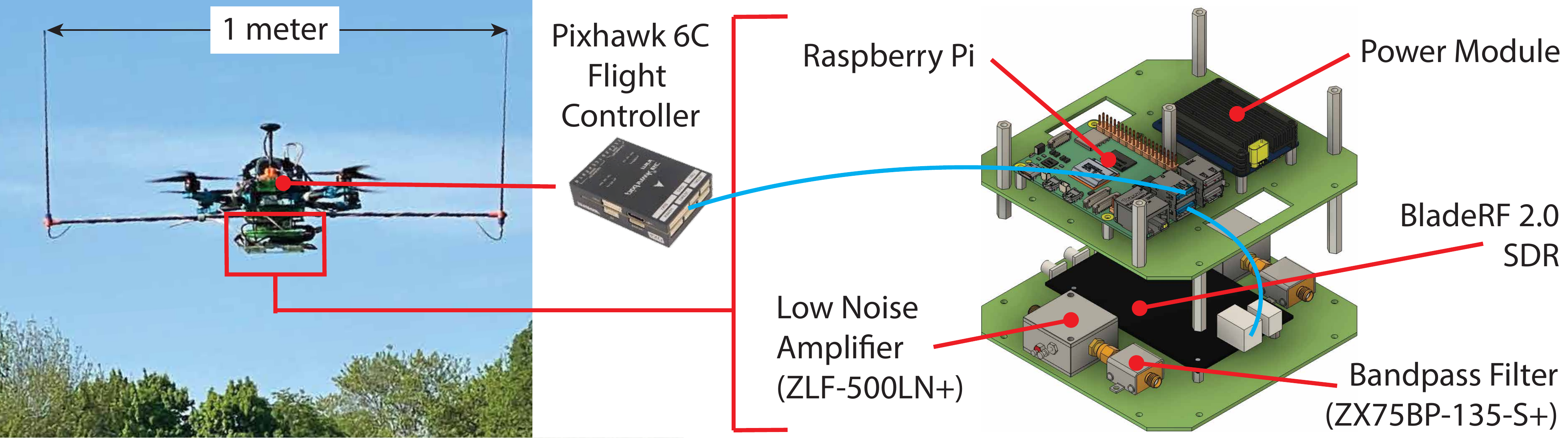}
  \vspace{-0.1in}
  \caption{\textbf{VHF sensing payload}. The sensing payload consists of a Raspberry Pi that interfaces with a software-defined radio and a PX4-based flight controller to collect raw VHF signals and UAV orientation for VHF AOA computation.}
  \label{fig:VHF_payload}
  \vspace{-0.15in}
\end{figure*}

Sperm whales stop vocalizing at the surface, making acoustic localization unreliable during this period. VHF-based sensing provides a complementary modality by enabling the localization of tagged whales when vocalizations are absent. Prior work has shown that VHF AOA can be obtained without directional ambiguity~\cite{scirob24}. We use a signal phase-based approach to compute VHF AOA to 100mW off-the-shelf fish trackers that transmit a pulse every second. We do so by having the UAV rotate in-place to emulate a virtual antenna array. Unlike received signal strength methods, which yield course AOA estimates using a directional antenna~\cite{Torabi2018UAVRTAS}, signal phase-based techniques can achieve much higher AOA accuracy~\cite{Arun2022P2SLAMBB,Gil2015AdaptiveCI}.
% Signal phase-based methods are capable of higher accuracy in AOA estimation in contrast to received signal strength based methods~\cite{Arun2022P2SLAMBB, Gil2015AdaptiveCI} which result in coarse AOA accuracy when using directional antennas~\cite{Torabi2018UAVRTAS}. 

The VHF sensing payload is comprised of a software-defined radio (SDR) and a Raspberry Pi (RPi) to collect and process VHF pulses using two omni-directional antennas. Low-noise amplifiers and bandpass filters are used to improve signal detection range (\Cref{fig:VHF_payload}). During data collection, the RPi receives phase data from the SDR via a SoapySDR API. A digital bandpass filter is applied iteratively to isolate pulses for each deployed tag frequency, and AOA is computed on-board using the WSR Toolbox~\cite{Jadhav2020AWS} (\Cref{fig:sample_filtered_signal_profile}). These AOA estimates are streamed to the state estimation module via Xbee.

\subsection{State Estimation Module}
The state estimation module estimates the belief locations for each vocalizing whale group, primarily using underwater acoustic AOA measurements, while addressing \textit{directional ambiguity} and \textit{group separation}. 
When a tagged whale has surfaced, VHF AOA estimates can supplement localization.
\subsubsection{Group Separation.}
The module begins by clustering acoustic AOA measurements, using a GMM to distinguish between multiple vocalizing groups. 
The GMM segments the received AOA estimates into $n$ clusters.
The number of clusters $n$ is chosen by minimizing the Bayesian information criterion, with $n \leq k$, where $k$ is a user-defined parameter for the total number of whales in the area.

% \noindent\textbf{Location Initialization.}
\subsubsection{Location Initialization.}
We use a particle filter to estimate belief locations. Several particles are sampled from the Gaussian distribution for each AOA cluster to initialize a new belief. Each belief consists of two candidate locations, one on each side of the sensor array, to account for the directional ambiguity inherent in the acoustic AOA estimates. The belief location particles are initialized using an approximate distance from the boat to the location where the whale was last seen diving. We apply particle filter updates to obtain new belief locations using the subsequent AOA measurements.

% \noindent\textbf{Sensor Maneuvering.}
\subsubsection{Sensor Maneuvering.}
The state estimation module can provide autonomous decisions for maneuvering the towed hydrophone array 
% At each maneuver decision interval, our system decides 
by either maintaining the current heading or taking a sharp turn towards the closest whale belief location. 
%We limit the number of times to take these maneuvers since we want to keep the hydrophone array horizontal to the ocean surface so that we can extrapolate its position relative to the boat without the use of an IMU. 
Periodic sharp turns help to triangulate the belief location, resolve directional ambiguity, and provide rough distance estimates. Additionally, we maintain proximity to the closest whale by assuming that the amplitude is inversely proportional to the source distance. If the amplitude of the received AOA for the nearest belief falls below a user-defined threshold, we head along the last received AOA direction.

% \noindent\textbf{AOA-Belief Matching and Surface Status.}
\subsubsection{AOA-Belief Matching and Surfacing Status.}
We first match the AOA clusters to the active belief locations using Hungarian matching. The matching minimizes the difference between the current AOA of the belief candidates to the hydrophone array and the mean of each of the AOA clusters. If a belief is unable to be matched to an observation for $\delta_{\text{silent}}$ consecutive seconds, or if a VHF AOA estimate is matched to the belief, the group is predicted to be on the surface. In the event of a sensor maneuver, we use the amplitude of the received signal to reject unlikely belief particles. We match AOA estimates before and after the maneuver using mean amplitude per GMM-estimated AOA candidate. For each belief, we eliminate belief location particles from the bimodal location distribution that do not match the newly estimated AOA after the maneuver.

\subsection{Autonomous UAV Control Module} 
Decision-making under uncertainty and partial observation is an area where RL offers significant potential~\cite{ralBhattacharya,SBTRO}. 
We extend our prior work~\cite{scirob24} on whale rendezvous to include realistic constraints like limited UAV flight times.
We formulate the real-time rendezvous as a partially observable sequential decision-making problem, where a belief $b =(b^1, \cdots, b^M)$ is represented as a collection of $M$ belief particles. 
Each belief particle $b^m$ consists of (i) the UAV's location and its remaining flight time, (ii) the whale's estimated location, and (iii) a possible whale surface schedule sampled from the Gaussian distributions of underwater and surface durations 
according to \cite{gero2014behavior}. 
The flight status $f \in \{\texttt{grounded}, \texttt{in-flight}\}$ determines the available actions $U_f$: $\{\texttt{takeoff},\texttt{wait}\}$ when \texttt{grounded}, or $\{\texttt{go-home},$ $\texttt{go-to-belief}\}$ when \texttt{in-flight}.
The UAV's navigation policy $\tilde{\mu}(b,f)$ at belief state $b$ chooses an action from all possible actions $U_f$ that minimizes the total time to rendezvous with a whale: 
% \begin{equation*}\vspace{-5pt}
$$\vspace{-5pt}\tilde{\mu}(b,f) \triangleq \argmin_{u \in U_f}\left\{
     \frac{1}{M} \sum_{m = 1}^M \big[ g(b^m,u) + J(F(b^m,u)) \big]\right\},$$
%      \vspace{-5pt}
% \end{equation*}
where $g(b^m, u)$ is the minimum of either the time taken to finish action $u$ for belief particle $b^m$, or a pre-defined number of time steps $\delta$. The transition function $F(b^m,u)$ gives the next belief particle given the current belief particle and action. 
This optimization procedure is similar to~\cite{scirob24}, but with the inclusion of a limited battery life constraint that uses a different cost function. The cost function $J(\cdot)$ is the time to complete rendezvous, considering both the remaining flight time of the UAV at maximum speed and the distance between the whale particle over the planning horizon.
Particularly, $J(b^m) = g(b^m,\bar{u}^m) + J(F(b^m, \bar{u}^m))$, where
the next action $\bar{u}^m$ for a \texttt{in-flight} UAV is set to \texttt{follow-whale} if enough battery is left to make a round-trip to the whale belief, otherwise \texttt{go-home}. 
The next action $\bar{u}^m$ for a \texttt{grounded} UAV is set to \texttt{takeoff} if rendezvous is possible in the next surfacing, otherwise \texttt{wait}. 
% \vspace{-5pt}
% \begin{equation*}
% \begin{aligned}
% &\begin{cases} \texttt{follow-whale}, & \text{if } \text{battery left to reach belief and return} \\
%     \texttt{go-home}, &\text{otherwise}\\
%     \end{cases} &\text{if UAV } \text{\texttt{in-flight}}\\
%     &\begin{cases}\texttt{takeoff}, &\text{if rendezvous is }\text{possible in the next surfacing}\\
%     \texttt{wait}, & \text{otherwise}\\
%     \end{cases}&\text{if UAV \texttt{grounded}}
%     \end{aligned}
% \end{equation*}

% \begin{equation*}
% \begin{aligned}
% \text{If  } = \texttt{in-flight},\\
%     &\bar{u}^m =\begin{cases} \texttt{follow-whale}, & \text{if } \text{enough battery is left to reach the belief and return} \\
%     \texttt{go-home}, &\text{otherwise}\\
%     \end{cases}\\
% \text{If  }&f_s =  \texttt{grounded},\\
% &\bar{u}^m =\begin{cases} 
%     \texttt{takeoff}, &\text{if the whale rendezvous is }\text{possible in the next surfacing}\\%, i.e.,} \\
%     %& \frac{||h-{w}^m||_2}{v_{max}} \leq \frac{\tau^m}{2} \text{ and } \delta_{\texttt{takeoff}} + \frac{||{w}^m - h||}{v_{max}}\leq b - t^m\\
%     % & \text{where }(a,b)\in I^m \text{ and } b\geq t^m\\
%     \texttt{wait}, & \text{otherwise}
%     \end{cases}
%     \end{aligned}
% \end{equation*}
% Here, \texttt{total-journey-time} 
% $=(||h-{w}^m||_2 + ||d^m-{w}^m||_2)/v_{max}$ 

\iffalse
\noindent The takeoff and navigation decisions are thus obtained by solving Eq~\ref{eq:drone_rollout}.
\fi
% \fi

% \subsection{Navigation Module}
Once the takeoff and navigation decisions are made with the RL-based approach, we send the commands to the UAV over a telemetry radio.
For rendezvous, we use a custom-built UAV with off-the-shelf Holybro S500 components and a Pixhawk flight controller. We use MAVProxy on the single MAVLink-based telemetry receiver to forward the packets from the UAV's telemetry radio to multiple devices to enable simultaneous autonomous control and monitoring. The UAV's FPV camera is only used for manual navigation by a human pilot and not for our autonomous rendezvous system.
\begin{table}[t]
    \centering
    \setlength{\extrarowheight}{1pt}
    \resizebox{\textwidth}{!}{  % Resizes the table to fit the text width
    \begin{tabular}{lccc}
    \toprule
    \makecell{Module and Configuration}
     & \makecell{Fielded (Dominica)\\ Experiments} 
     & \makecell{Land-based\\ Experiments} 
     & \makecell{Simulation}\\
    \midrule
    \textbf{Sensing}: Acoustic AOA & \textcolor{OliveGreen}{\checkmark}  & N/A  & \textcolor{OliveGreen}{\checkmark} \\
    \textbf{Sensing}: VHF Pulse Detection                   & \textcolor{OliveGreen}{\checkmark} & \textcolor{OliveGreen}{\checkmark}  & N/A  \\
    \textbf{Sensing}: VHF AOA
     & \textcolor{red}{$\times$} & \textcolor{OliveGreen}{\checkmark}  & \textcolor{OliveGreen}{\checkmark}  \\
    \textbf{State Estimation}: Boat maneuver decisions      & Manual & N/A & Autonomous \\
    \textbf{UAV Control}: Belief candidate selection & Manual & N/A & Autonomous \\
    \textbf{UAV Control}: Takeoff decisions & Manual & N/A & Autonomous \\
    \textbf{UAV Control}: Navigation & Autonomous & N/A & Autonomous \\
    \bottomrule
\end{tabular}}
\vspace{0.05cm}
\caption{System configurations tested in hardware and simulation experiments.}
\vspace{-0.85cm}
\label{tab:comparison_table}
\end{table}

\iffalse
\begin{figure}[!p]
    \begin{subfigure}{0.95\textwidth}
        \centering
        \includegraphics[scale=0.43]{Figures/ISER_2025_RDV1.pdf}
        % \caption{\textbf{First In Situ Rendezvous.} First successful real-time rendezvous attempt in Dominica. (A) Acoustic AOA estimates from multiple groups of whales are clustered using a Gaussian Mixture Model. (B) UAV trajectory from takeoff to the belief location. (C) Snapshots of the rendezvous attempt taken from the UAV's FPV camera.}
        \label{fig:RSS_2025_Sperm_whale_rendezvous_1}
    \end{subfigure}
    \begin{subfigure}{0.95\textwidth}
        \centering
        \includegraphics[scale=0.43]{Figures/ISER_2025_RDV2.pdf}
        % \caption{\textbf{Second In Situ Rendezvous.} Second successful real-time rendezvous attempt in Dominica showing (A) Acoustic AOA estimates, (B) UAV trajectory, and (C) Snapshots from the UAV's FPV camera.}
        \label{fig:RSS_2025_Sperm_whale_rendezvous_2}
    \end{subfigure}
    \caption{\textcolor{red}{\textbf{First Rendezvous.} First successful real-time rendezvous attempt in Dominica. (A) Acoustic AOA estimates from multiple groups of whales are clustered using a Gaussian Mixture Model. (B) UAV trajectory from takeoff to the belief location. (C) Snapshots of the rendezvous attempt taken from the UAV's FPV camera. \textbf{Second Rendezvous.} Second successful real-time rendezvous attempt in Dominica showing (A) Acoustic AOA estimates, (B) UAV trajectory, and (C) Snapshots from the UAV's FPV camera.}}
    \label{fig:ISER_2025_rendezvous}
\end{figure}
\fi

\begin{figure}[!p]
    \includegraphics[scale=0.38]{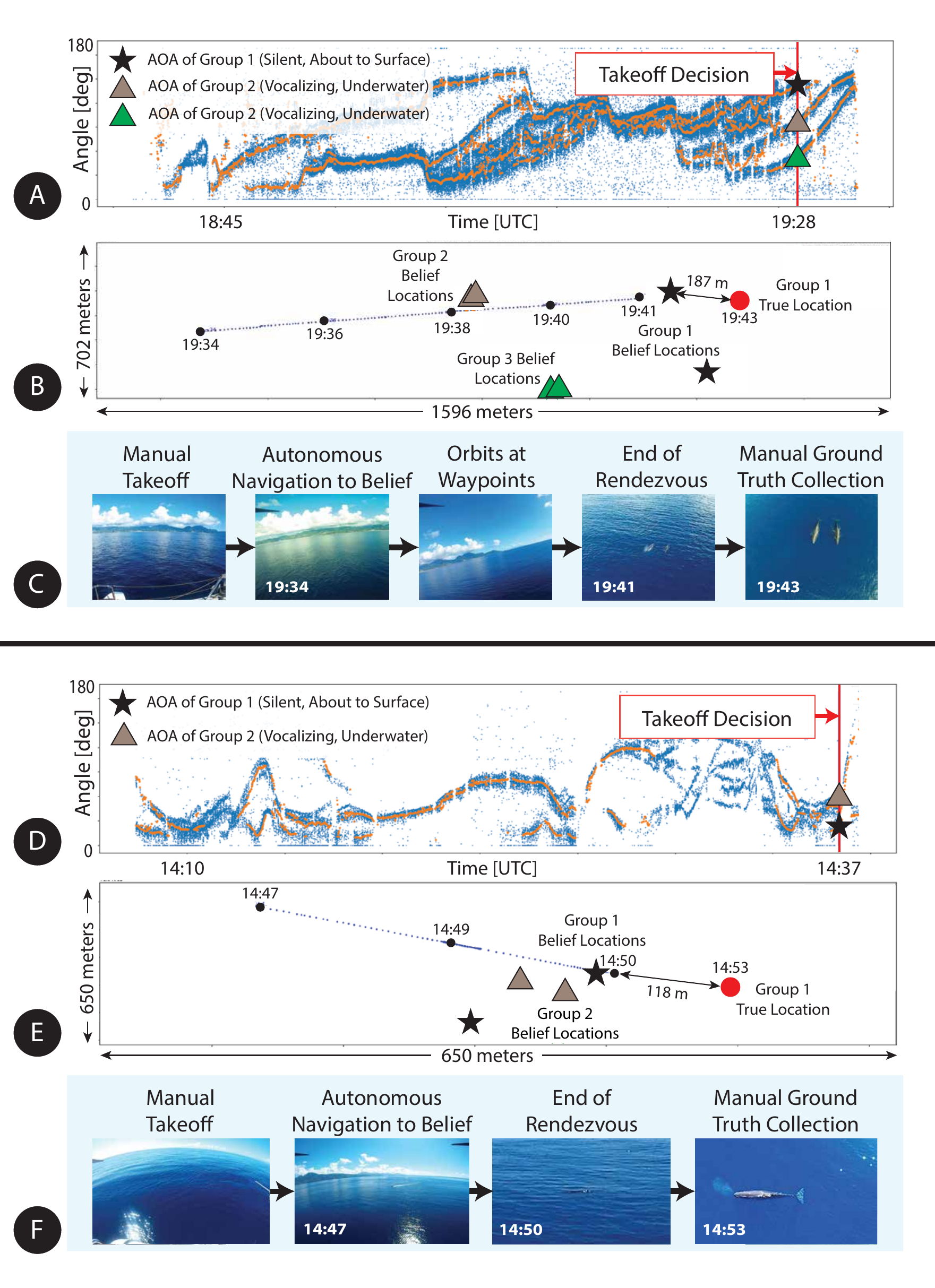}
    \caption{\textcolor{black}{\textbf{In situ whale rendezvous.} First (A, B, C) and second rendezvous attempts (D, E, F) in Dominica. The stars indicate the group that stopped vocalizing, while triangles indicate groups that continue to vocalize. (A, D) Acoustic AOA estimates from multiple groups of whales are clustered using a Gaussian Mixture Model. (B, E) UAV's autonomous navigation trajectory. (C, F) Snapshots of the rendezvous attempt taken from the UAV's FPV camera. 
    % \textbf{Second Rendezvous.} Second successful real-time rendezvous attempt in Dominica showing (D) Acoustic AOA estimates, (E) UAV trajectory, and (F) Snapshots from the UAV's FPV camera.
    }}
    \label{fig:ISER_2025_rendezvous}
\end{figure}

% tot_drone_dist= 891.3500936367769 
% Errors: gt_drone_distance: 184.04716999414413 
% gt_bel0_distance: {0: np.float64(151.0866093785069), 1: np.float64(519.1755240532042), 2: np.float64(528.3718146208404)} 
% gt_bel1_distance: {0: np.float64(340.72955546310874), 1: np.float64(510.2542427526399), 2: np.float64(520.9258040929203)} 
% drone_bel0_distance: {0: np.float64(34.27310240413004), 1: np.float64(339.99780846696484), 2: np.float64(458.1771766830671)} 
% drone_bel1_distance: {0: np.float64(400.6438601904477), 1: np.float64(330.6701942446299), 2: np.float64(451.9475957106307)} 

% tot_drone_dist= 401.17083948564715 
% Errors: gt_drone_distance: 117.51454419651233 
% gt_bel0_distance: {0: np.float64(263.76984389088216), 1: np.float64(147.44596860207517)} 
% gt_bel1_distance: {0: np.float64(117.5665687097451), 1: np.float64(119.11057585271912)} 
% drone_bel0_distance: {0: np.float64(234.32389358750237), 1: np.float64(74.64205508385199)} 
% drone_bel1_distance: {0: np.float64(0.058444787677106855), 1: np.float64(77.78314271149205)} 

\vspace{-1pt}
\section{Results} \label{results}
We performed in situ experiments in the Caribbean Sea along the western coast of Dominica to validate different aspects of our system. We also report land-based VHF tests in hardware, and a sensor-noise sensitivity analysis of localization error and evaluation of the decision-making policy in simulation.
% \Cref{tab:comparison_table} summarizes the configurations tested across different environments. 

% \subsection{Dominica Experiments}

\subsection{Rendezvous Experiments with Acoustic AOA in Dominica}
% We present proof-of-concept experiments in Dominica, where we achieved rendezvous within 200 meters away from a surfaced sperm whale on two different occasions. 

\subsubsection{Setup.} 
% In this experiment, we used acoustic AOA to track whales and used a UAV to autonomously navigate to whale locations when they were at surface. We took a few manual decisions (~\Cref{tab:comparison_table}) of acoustic sensor maneuvers, take off times and belief selection.
% During the whales' underwater phase, we manually maneuvered the hydrophone array approximately every ten minutes to resolve directional ambiguity in the acoustic AOA estimates. During a rendezvous attempt, a pilot manually performed UAV take-off and landing. After the UAV reaches an altitude of 30 meters, the UAV control module took over and navigated the UAV to the belief location. We report two out of five total rendezvous attempts for which we had true whale positions; this was done by flying directly above the whale. The drone pilot continuously monitored the FPV video stream during the rendezvous attempt for safety purposes. We note that our system does not currently integrate any visual observations. The UAV performed orbits at each waypoint so the pilot can visually scan for whales for obtaining ground truth locations. Once surfaced whales were clearly sighted, the pilot took over manual control to collect ground-truth. When belief particles did not converge, the UAV moved towards the nearest surfaced whale belief candidate. This way of selecting the belief candidate was arbitrary in the Dominica experiments, which we further improve in the simulation study with automated belief selection.

In this experiment, acoustic AOA was used to track whales, and a UAV autonomously navigated to whale locations at the surface, with manual decisions made for sensor maneuvers, UAV takeoff timing, and belief candidate selection (\Cref{tab:comparison_table}). When the whales were underwater, manual decisions were made to maneuver the hydrophone array approximately every ten minutes to resolve directional ambiguities. 
We limit the number of times to take these maneuvers since we want to keep the hydrophone array horizontal to the ocean surface so that we can extrapolate its position relative to the boat since the towed array did not have an IMU. 
For each rendezvous attempt, UAV takeoff and landing were manually controlled, but once the UAV reached 30 meters altitude, it autonomously navigated to the belief location. We report two out of five total rendezvous attempts since these are the ones for which we had true whale positions by flying directly above the whale. The drone pilot monitored the FPV video feed throughout for surfaced whales. Visual observations were not integrated into the system; instead, the UAV orbited at each waypoint to allow the pilot to visually identify surfaced whales for obtaining ground-truth. Upon a clear visual confirmation, manual control was used to collect ground-truth data. When particles failed to converge under directional ambiguity, the UAV navigated to the nearest surfaced-whale belief candidate — a heuristic used in the Dominica trials and later replaced by automated candidate selection (see Sec~\ref{sec:simulation-setup}) in simulation.

\subsubsection{Rendezvous Attempt 1 (\Cref{fig:ISER_2025_rendezvous} A-C).} We tracked three distinct vocalizing whale groups. Each group consisted of multiple whales moving in close proximity to one another. We initialized three sets of beliefs and localized the groups using acoustic AOA for a total duration of 52 minutes over a distance of over $5.2$ km. 
During the rendezvous attempt, we chose to navigate our UAV to the belief corresponding to the first whale group that stopped vocalizing for one minute. 
% After launching the UAV, it autonomously navigated towards the belief location.
We stopped autonomous control approximately $34$ meters from the belief location because the pilot could clearly detect the whales visually in the video stream. 
The UAV pilot manually guided the UAV to be directly above the whales to capture their ground-truth position. The total duration of the autonomous flight was approximately eight minutes, during which the UAV covered a distance of about $891$ meters. This attempt had a localization error of $186.56$ meters.

\subsubsection{Rendezvous Attempt 2 (\Cref{fig:ISER_2025_rendezvous} D-F).} 
We tracked two whale groups for a duration of $30$ minutes over $2$ km. 
% We followed a similar procedure to the first rendezvous attempt. 
The autonomous flight lasted for $3$ minutes,
% The total duration of the autonomous flight was approximately $3$ minutes, 
during which the UAV covered $401$ meters till it reached the belief location, which was $117.5$ meters away from the ground-truth whale position. After the UAV reached the belief location, the drone pilot took ground-truth positions using manual control. 

\begin{figure*}[t]
\centering
\includegraphics[scale=0.375]{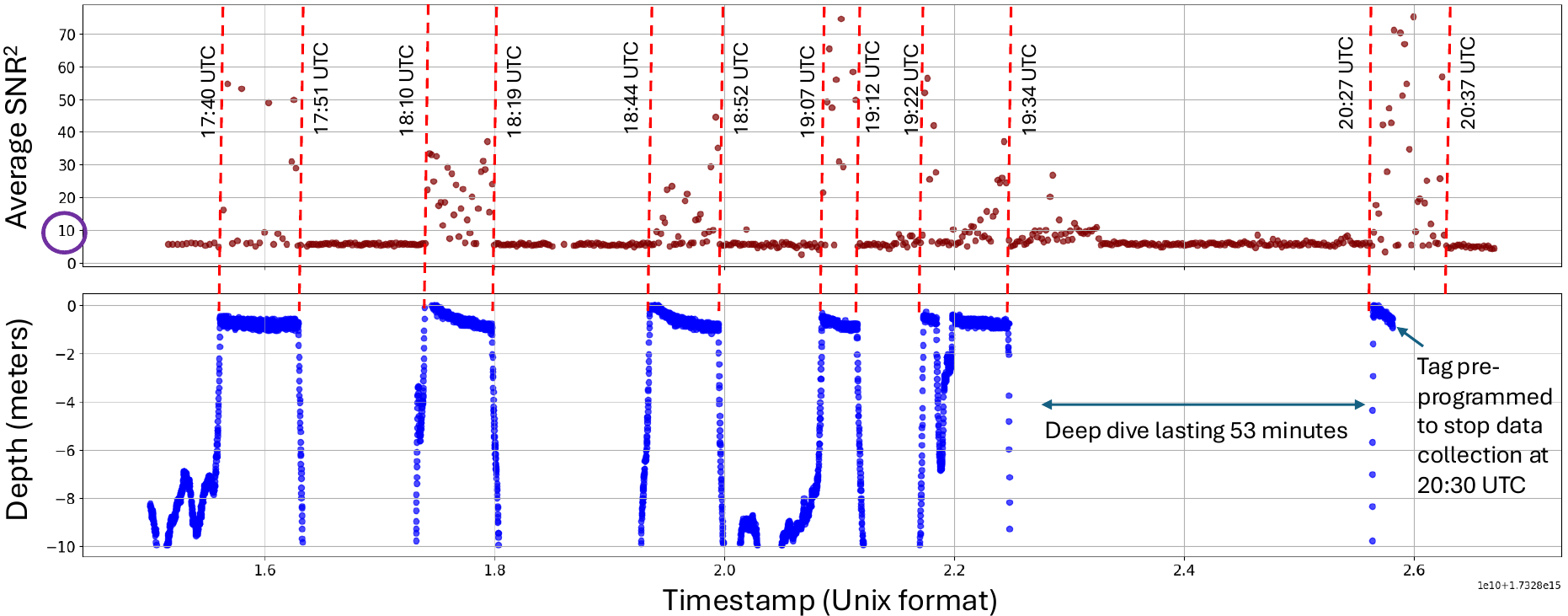}
\vspace{-0.075in}
\caption{\textbf{VHF Signal Detection for a Tagged Whale}. VHF signal detection time vs. data from the tag depth sensor. We find that instantaneous detection of whale surfacing events is possible. The whale dived up to 800 meters; is clipped in the bottom plot.}
\vspace{-0.2in}
\label{fig:vhf_ping_detection_from_tag}
\end{figure*}

\begin{figure*}[t]
    \centering
    \includegraphics[scale=0.19]{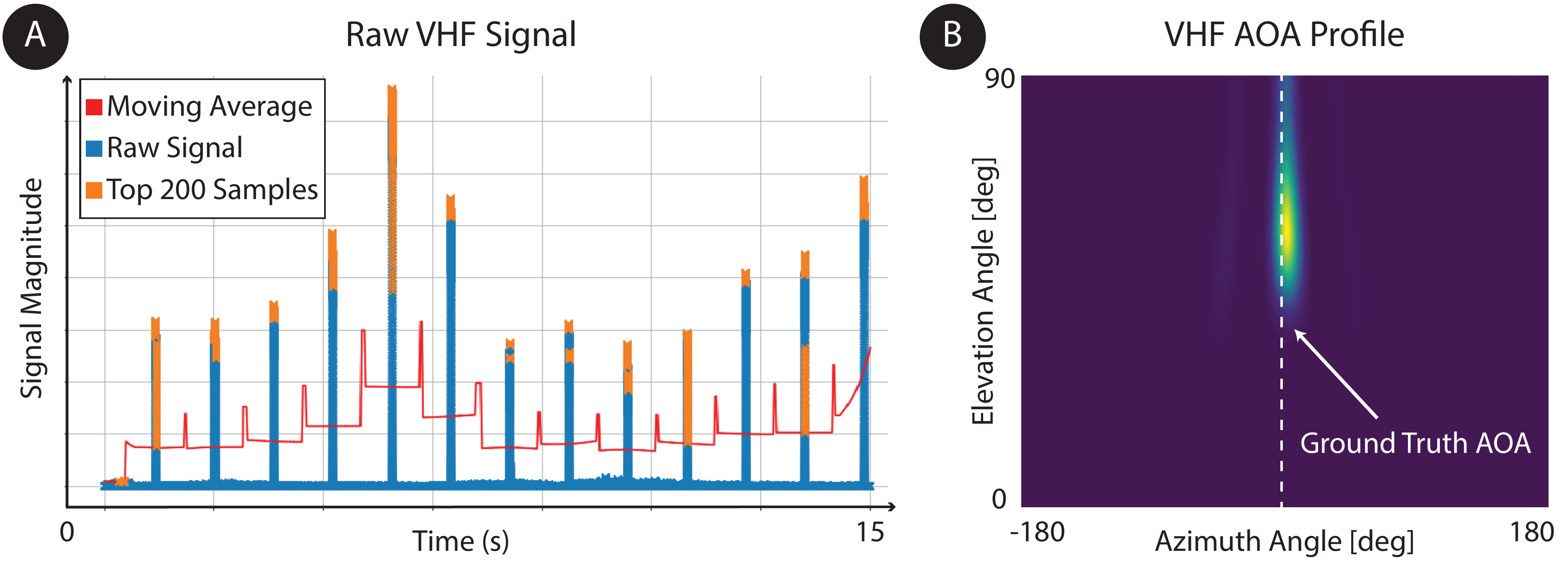}
    \vspace{-0.25in}
    \caption{\textbf{VHF AOA computation.} (A) Raw signal obtained from a VHF tag, which is filtered using a moving average. The top 200 samples for each pulse are used for AOA computation. (B) The computed AOA profile shows a clear peak around zero degrees, close to the ground truth AOA.}
    \vspace{-0.2in}
    \label{fig:sample_filtered_signal_profile}
    \vspace{0.2cm}
\end{figure*}

\subsection{VHF Pulse Detection Dominica}
We performed experiments to validate the use of VHF pulses to detect whale surfacing events and the detection range of our sensing payload. The results show that the pulses reliably indicated when a tagged whale surfaced and were detectable from up to two kilometers away.
% We find that these pulses are a good indicator that a tagged whale has surfaced and are able to detect these pulses from two kilometers away. 
We used the F1800 fishtracker by ATSTRACK, that emits a pulse at a fixed VHF frequency for 20 ms every 1100 ms. 
% The pulses were collected via a marine VHF omni-directional antenna mounted on the catamaran's mast, 10 meters above the water surface. 
A marine VHF omni-directional antenna mounted 10 meters above the water on the catamaran's mast was used as a receiver.

\subsubsection{Setup (Surfacing Detection).} We first tested the VHF signal detection to a tagged whale using the tag developed by Project CETI~\cite{iScienceCETI}. We collected data for multiple surfacings over three hours. We use the tag's pressure sensor data to cross-validate the signal detection with whale surfacing events. These results show that instantaneous detection of a tagged whale's surfacing event is possible~(\Cref{fig:vhf_ping_detection_from_tag}). We empirically chose a threshold of 10 for the mean square signal-to-noise ratio. The variance of SNR$^2$ can be attributed to the bobbing motion of the surfaced whale, with waves washing over the tag intermittently, obscuring the signal.

\begin{table}[t]
    \centering
    \setlength{\extrarowheight}{1pt}
    \makebox[\textwidth][c]{%
        \begin{tabular}{@{}l@{\hskip 1em}c@{\hskip 1em}c@{\hskip 2em}c@{\hskip 1em}c@{}}
        \toprule
        \multirow{2}{*}{\makecell[l]{\textbf{Fish-Tracker} \\ \textbf{Freq. (MHz)}}}
         & \multicolumn{2}{c}{\textbf{Antenna Separation: 1 m}} 
         & \multicolumn{2}{c}{\textbf{Antenna Separation: 2 m}} \\
        \cmidrule(lr){2-3} \cmidrule(lr){4-5}
         & Mean SNR & Avg. Min. Error & Mean SNR & Avg. Min. Error \\
        \midrule
        $150.453$ & $26.48$ dB & $18.07^\circ \pm 8.28^\circ$ & $23.16$ dB & $8.85^\circ \pm 3.64^\circ$ \\
        $150.754$ & $29.40$ dB & $6.04^\circ \pm 1.67^\circ$ & $27.57$ dB & $2.27^\circ \pm 1.19^\circ$ \\
        $150.983$ & $24.15$ dB & $14.09^\circ \pm 9.86^\circ$ & $21.08$ dB & $5.64^\circ \pm 1.79^\circ$ \\
        \bottomrule
        \end{tabular}
    }
    \vspace{0.1cm}
    \caption{Minimum VHF AOA estimation error for the top three angles in the AOA profile averaged over five samples per combination in land-based experiments.}
    \vspace{-0.3in}
    \label{tab:vhf_aoa_error}
\end{table}
\begin{table}[t]
    \centering
    \begin{tabular}{l@{\hskip 1em}c@{\hskip 1em}c}
    \toprule
    \multirow{2}{*}{\bfseries Sensor Noise Std. Dev.} & \multicolumn{2}{c}{\makecell{\textbf{Average Minimum} \textbf{Relative Localization Error}}} \\
        \cmidrule(lr){2-3}
     & Acoustic Only & Acoustic + VHF \\
    \midrule
    $0.1^\circ$ & $1.00 \pm 0.25$ & $0.23 \pm 0.11$ \\
    $5^\circ$   & $0.90 \pm 0.19$ & $0.37 \pm 0.16$ \\
    $10^\circ$  & $1.24 \pm 0.26$ & $0.39 \pm 0.15$ \\
    $15^\circ$  & $1.54 \pm 0.44$ & $0.25 \pm 0.12$ \\
    \bottomrule
    \end{tabular}
    \vspace{0.1 cm}
    \caption{Localization error with varying noise conditions for acoustic-only and acoustic + VHF AOA from zero mean Gaussian distributions with the same standard deviation in simulation. The relative localization error is reported as the 99\% confidence interval for the minimum localization error over the surface duration, averaged over $75$ surfacings
    % average minimum localization error 
    relative to the performance of acoustic-only localization with $0.1^\circ$ standard deviation.
    }
    \label{tab:sensitivity_analysis_sensor_noise}
    \vspace{-0.325in}
\end{table}

\begin{wrapfigure}[16]{r}{0.525\textwidth}
\vspace{-0.45in}
\centering
\includegraphics[scale=0.2]{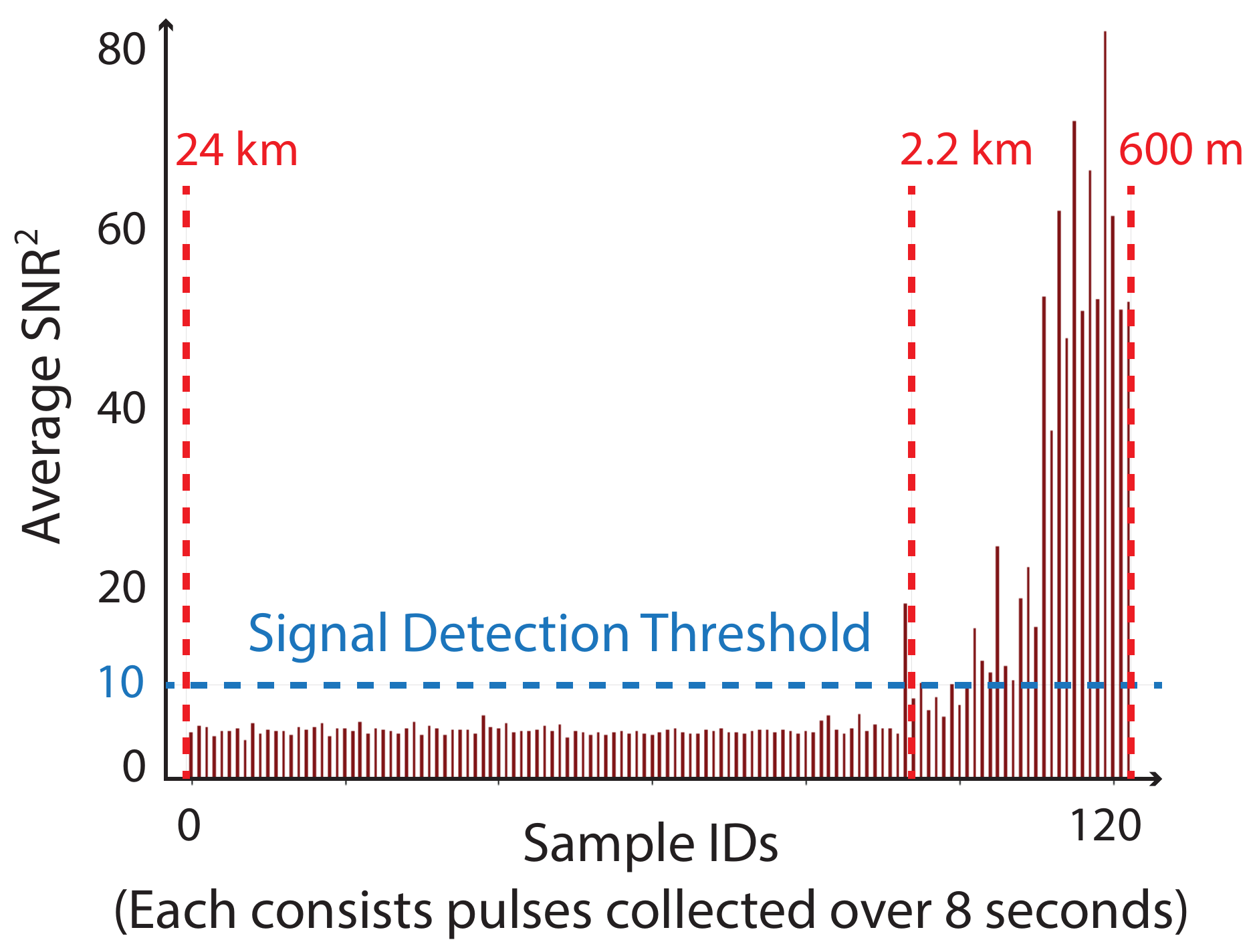}\vspace{-0.11in}
\caption{\textbf{Range of VHF pulse detection}. 
The signal was detected with a sampling interval of 1 minute when the tag, floating in the water near the dock, was within 2 km of the antenna mounted on the catamaran's mast.
% The tag, floating in water near the mooring at the dock, was sampled every 1 minute. The VHF signal was successfully detected when the tag was within 2 kilometers of the antenna mounted on the catamaran’s mast.
% The tag was floating in the water near the dock. 
% The time between consecutive samples is 1 minute. 
% The sampling interval was 1 minute.
}
\label{fig:vhf_ping_detection_range_test}
% \vspace{-0.3in}
\end{wrapfigure}
\subsubsection{Setup (Detection Range).} We then conducted a signal detection range test to a tag deployed in water at the catamaran's mooring near the dock (\Cref{fig:vhf_ping_detection_range_test}). We use the open source code provided by~\cite{scirob24} to collect data for these experiments. We use a Fast Fourier Transform for signal detection, setting the spectral power threshold to $-77$ dBm. We find that the integration of the low-noise amplifiers greatly increases our signal detection range. Clear signal detection was observed within a range of two kilometers. As illustrated in \Cref{fig:vhf_ping_detection_range_test}, the SNR improved significantly when the distance to the tag was less than one kilometer away, demonstrating the setup's effectiveness in identifying surfacing events at this distance. This is particularly relevant, as seen in our fielded experiments where the whales surfaced within a kilometer of the catamaran's location.

\subsection{Land-based VHF AOA experiments}
The VHF experiments in Dominica show the potential of detecting VHF pulses to a tagged whale and its potential integration as a signal for surfacing detection. Here, we show the ability to compute VHF AOA to collected pulses, which helps aid localization for surfaced, non-vocalizing whales.

\subsubsection{Setup.} Hardware experiments on land were conducted separately to test VHF AOA estimation to multiple fish-trackers using on board computation with the UAV payload.
We validated the performance of the VHF sensing payload for AOA computation at Harvard's Ohiri field. During the experiments, we flew the UAV at an altitude of seven meters with three fish-trackers elevated at a height of one meter from the ground, approximately equally spaced on a 60-meter radius circle. We use an antenna separation of one and two meters to characterize the impact of antenna separation on performance. 

Fig.~\ref{fig:sample_filtered_signal_profile} shows a sample output for AOA estimation to a fish-tracker with two meter separation between the UAV antennas collecting data over 15 seconds. Table~\ref{tab:vhf_aoa_error} shows the minimum error in AOA estimation over the top three angles in the AOA profile, for one and two meter antenna separations, averaged over $5$ samples each.

% \begin{figure*}[t]
% \vspace{-0.15in} \centering \includegraphics[scale=0.35]{Figures/ISER_2025_End_to_End_demonstration.pdf} 
% \vspace{-0.1in} 
% \caption{\textbf{End-to-End autonomous demonstration}. Shows real-time decision-making module and data acquisition with ground robots emulating engineered whales and GPS AOA in lieu of acoustic AOA. We separately demonstrate a proof-of-concept VHF AOA computation using a UAV with onboard sensing and computation.} \label{fig:end_to_end_exp_overview} 
% \vspace{-0.1in} 
% \end{figure*} 

% \begin{wrapfigure}{r}{0.55\textwidth}
% \begin{figure*}[t]
% \centering
% % \vspace{-0.325 in}
% \includegraphics[scale=0.25]{Figures/Temp_image.png}
% \vspace{-0.175in}
% \caption{Accuracy results for AOA experiments on land with all onboard sensing and computation.}
% \label{fig:vhf_aoa_accuracy}
% % \end{wrapfigure}
% \end{figure*}
% \begin{table}[h!]
% \centering
% \begin{tabularx}{\textwidth}{|X|X|X|}
% \hline
%  & \textbf{Acoustic AOA} & \textbf{Acoustic and VHF AOA} \\
% \hline
% $\sigma_\text{acoustic} = 5^\circ$ & $150.65$ & Row 1, Col 3 \\
% \hline
% $\sigma_\text{acoustic} = 10^\circ$ & $224.74$ & Row 2, Col 3 \\
% \hline
% $\sigma_\text{acoustic} = 20^\circ$ & $458.18$ & Row 3, Col 3 \\
% \hline
% \end{tabularx}
% \caption{Ablation study}
% \end{table}
% \begin{wrapfigure}{r}[0pt]{0.35\textwidth}

% \subsection{Effect of Sensor Noise on Localization}
\subsection{Simulation Study}
Here, we show how sensor noise affects state estimation. We show the system's capability of automating components not validated in the Dominica experiments, such as sensor maneuvers, takeoff decisions, and belief candidate selection (\Cref{tab:comparison_table}).

% In this section, we show how is state estimation process gets affected by various sensor noises. We also show the system's capability of achieving autonomous rendezvous, where we automate various aspect of the system, which weren't validated in the Dominica experiments, such as autonomous sensor maneuvers, takeoff decisions and belief selection (See~\Cref{tab:comparison_table}).
\begin{wrapfigure}[23]{r}[0pt]{0.65\textwidth}
    \centering
    \vspace{-0.15in}
    \includegraphics[scale=0.5]{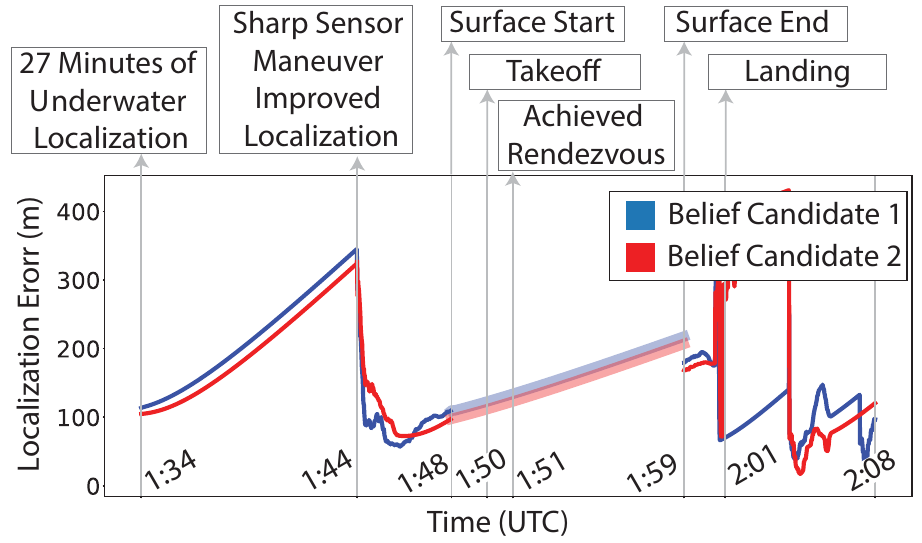}\vspace{-10pt}
    \caption{\textbf{Decision-making snapshot.} Shows belief localization errors overlayed with the system's autonomous decisions for a simulation experiment. The whale trajectory is obtained by interpolating visual observations and depth information from on-body tags~\cite{dswp}. We simulate acoustic AOA with Gaussian error of std. dev. $5^\circ$. The state estimation module autonomously decides to maneuver the hydrophone array to eliminate directional ambiguity. The UAV flew for 10 minutes, achieving rendezvous within 118 meters of the whale.}
    \label{fig:Decision_making_plot}
    % \vspace{-0.8cm}
\end{wrapfigure}

\subsubsection{Setup.} \label{sec:simulation-setup}
The DSWP dataset~\cite{dswp} was used to simulate whale trajectories over three consecutive surfacings. Dive schedules were derived from depth data on on-body tags, and trajectories were interpolated from visual encounters. Sensor measurements were simulated by adding Gaussian noise to the true angle-of-arrival (AOA) from the boat location. To replicate the catamaran's speeds, the simulated boat operated within a speed range of 2--6 m/s and made autonomous maneuver decisions every 5 minutes. UAV takeoff and landing decisions were also autonomously executed. 
% The UAV control system attempted to rendezvous with the nearest group's belief. 
The UAV control module selected the closest surfacing belief candidate and, upon arrival, waited for 1 minute before proceeding to the next candidate. UAV behavior was simulated using PX4's Software-In-The-Loop (SITL) environment with a max drone speed of $10$ m/sec.

\subsubsection{Effect of sensor noise on state estimation.}
~\Cref{tab:sensitivity_analysis_sensor_noise} evaluates localization error under zero-mean Gaussian sensor noise with standard deviations of
$\{0.1^\circ, 5^\circ, 10^\circ, 15^\circ\}$. 
For each sensor setup and across three surfacing events, 25 trials were conducted, recording the minimum localization error during each surfacing and averaging these values. The results were normalized against the average error from acoustic AOA-only localization at $0.1^\circ$ noise.
% For each sensor combination and each of the three surfacing events, we ran 25 trials, recorded the minimum localization error over the whale’s entire surfacing duration, and averaged those minimum errors. They are then normalized by the average error of acoustic AOA-only localization with $0.1^\circ$ standard deviation.
We note that with acoustic AOA-only localization, the minimum error typically occurs around the first second, while when integrating VHF AOA, the minimum error occurs approximately 230 seconds into the surfacing duration. We verify that localization error increases as sensor noise increases. In some cases, we find that acoustic AOA-only localization suffers from belief re-initializations before surfacing, an issue that is mitigated by having VHF AOA estimates on the surface. The localization improvement with VHF AOA can also be attributed to its lack of directional ambiguity.

% We note that for acoustic AOA only localization this is typically the first second after the whale surfaces. They are then normalized by the average performance of acoustic AOA only localization with a sensor of standard deviation of 0.1 degrees. 
% We verify that the localization error increases as sensor noise increases when using only acoustic AOA. We also find that the same trend emerges when integrating VHF AOA
% % , except that the overall localization error performance is far better
% % since the UAV is able to re-route itself 
% as new VHF AOA estimates are incorporated into the location estimate. 
% In general, we find that acoustic AOA-only localization suffers from poorly timed belief re-initializations, an issue that is mitigated from having available VHF AOA estimates on the surface. We also note that the minimium relative localization accuracy is achieved when integrating VHF AOA approximately 230 seconds into the surfacing duration, irrespective of sensor noise. This poses a tradeoff between timely rendezvous and improved localization.

% \textcolor{red}{DONE WRITING UP TILL HERE}

% \subsection{Decision-Making Study}
% We perform simulation experiment with a whale trace from DSWP dataset. 
% The whale was tracked over $41$ minutes of underwater phase with acoustic AOA with an error of $\sigma_{acoustic}=5$ degree. 
\subsubsection{Autonomous decision-making.}
\Cref{fig:Decision_making_plot} shows the autonomous decisions taken at different points of the UAV navigation. 
% With a $200$-meter rendezvous radius, this snapshot achieved rendezvous success. 
We consider rendezvous to be successful when the UAV reaches within $200$ meters of a surfaced whale.
We find that the rendezvous success increases with increasing flight time. We obtain $91.5, 96.6$, and $98.3$ percent success rates with $5, 10$, and $15$ minutes of flight times, respectively, averaged over $40$ runs with whale traces from the DSWP dataset.
\section{Conclusion}
\label{conclusion}
We present a real-time whale-rendezvous system using in situ sensing, advancing sperm-whale tracking. Field trials off Dominica validate state estimation, autonomous UAV control, and VHF pulse detection with some components (sensor maneuvers, belief-candidate selection, takeoff) being manually guided. Simulations and land-based tests covered autonomous features not exercised at sea and VHF AOA estimation, respectively. Our main contribution is integrating biological and operational priors (dive patterns, maneuver strategies) with existing methods—particle filters, Gaussian mixture models, synthetic aperture radar, and reinforcement learning. Future work targets distributed UAV-borne VHF sensing for improved surface tracking and multi-whale rendezvous via networked UAVs with onboard sensing and processing.

\bibliographystyle{styles/bibtex/spmpsci}
\bibliography{main}

\end{document}